\documentclass[10pt,twocolumn,letterpaper]{article}

\usepackage[pagenumbers]{cvpr} 

\usepackage{bbding}
\usepackage{multirow}
\usepackage{multicol}
\usepackage[table]{xcolor}
\usepackage{float}
\usepackage[export]{adjustbox}
\usepackage[utf8]{inputenc}
\usepackage{graphicx}
\usepackage{amsmath}
\usepackage{amssymb}
\usepackage{booktabs}
\usepackage{stackengine}
\setstackgap{S}{2pt}
\setstackgap{L}{.4\baselineskip}
\usepackage[pagebackref,breaklinks,colorlinks]{hyperref}

\usepackage[capitalize]{cleveref}
\crefname{section}{Sec.}{Secs.}
\Crefname{section}{Section}{Sections}
\Crefname{table}{Table}{Tables}
\crefname{table}{Tab.}{Tabs.}

\def\cvprPaperID{*****} 
\def\confName{CVPR}
\def\confYear{2023}

\begin{document}

\title{MSANet: Multi-Similarity and Attention Guidance for Boosting Few-Shot Segmentation}

\author{Ehtesham Iqbal$^*$  \qquad Sirojbek Safarov$^*$ \qquad Seongdeok Bang$^\dagger$ \\
AiV Research Group,\;South Korea\\
{\tt\small iqbal.ehtesham@aiv.ai}\\
{\tt\small safarov.sirojbek@aiv.ai}\\
{\tt\small bang.seongdeok@aiv.ai}}

\maketitle
\begin{NoHyper}
\def\thefootnote{*}\footnotetext{Equal Contribution.}
\def\thefootnote{$\dagger$}\footnotetext{Corresponding Author.}
\end{NoHyper}
\def\thefootnote{\arabic{footnote}}
\begin{abstract}
Few-shot segmentation aims to segment unseen-class objects given only a handful of densely labeled samples. Prototype learning, where the feature extracted from support images yields a single or several prototypes by averaging global and local object information, has been widely used in FSS. However, utilizing only prototype vectors may be insufficient to represent the features for all support images. To extract abundant features and make more precise predictions, we propose a \textbf{M}ulti-\textbf{S}imilarity and \textbf{A}ttention \textbf{N}etwork (MSANet) including two novel modules, a multi-similarity module and an attention module. The multi-similarity module exploits multiple feature-maps of support images and query images to estimate accurate semantic relationships. The attention module instructs the MSANet to concentrate on class-relevant information. The network is tested on standard FSS datasets, PASCAL-$5^i$ 1-shot, PASCAL-$5^i$ 5-shot, COCO-$20^i$ 1-shot, and COCO-$20^i$ 5-shot. The MSANet with the backbone of ResNet101 achieves the state-of-the-art performances for all 4-benchmark datasets with mean intersection over union (mIoU) of 69.13\%, 73.99\%, 51.09\%, 56.80\%, respectively. Code is available at \url{https://github.com/AIVResearch/MSANet}. 
\end{abstract}
\begin{figure}[t]
  \centering
 \includegraphics[width=1\linewidth,height=1\linewidth]{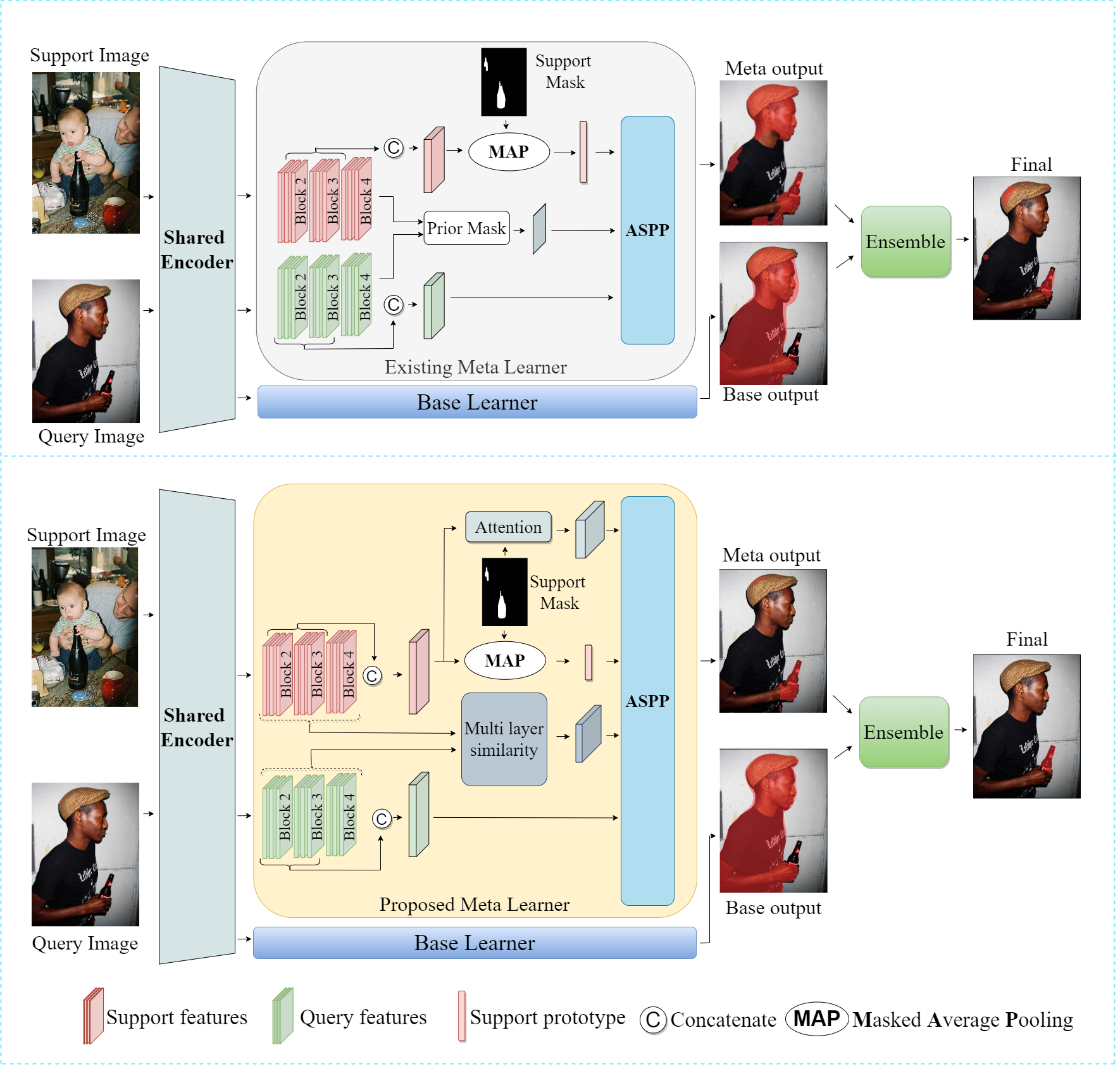}
\caption{Comparison a meta learner between the existing network and the MSANet. The main difference is that the former uses only class-representative prototype vectors, while the MSANet includes the multi-similarity module for visual correspondences and an attention module for target category focus. The rest of the network is the same as the architecture of BAM \cite{BAM}}.
\label{fig:first}
\end{figure} 
\section{Introduction}
\label{sec:intro}
Following the development of well-established large-scale datasets \cite{large,coco,pascal,SDS}, a series of supervised convolutional neural networks (CNNs) have shown great potential for semantic segmentation tasks \cite{sem1,sem2,sem3,sem4,accvsem4}. The performance of these supervised CNNs is highly dependent on the quality and quantity of training datasets such as the numbers of well-annotated data, the balance of class distribution, and sample representation. However, in real-world applications, it is difficult to secure a lot of annotated data, especially in dense prediction tasks \cite{inst1,inst2,inst3,inst4,inst5,accvinst6}. Moreover, traditional supervised CNNs may struggle with generalization capability on the images with unseen classes.

Inspired by the human cognitive ability to distinguish objects with only a few input data, a few-shot learning (FSL) technique is developed \cite{FSL1,FSL2,FSL3,FSL4}. This technique builds a network that can be generalized to unseen domains with few available annotated samples. Few-shot segmentation (FSS) \cite{FSS1shaban,FSS2PFE,FSS3,FSS4,FSS5,FSS6crnet,FSS7,FSS8,FSS9Hsnet,FSS10,FSS11,FSS12,FSS13PMM,FSS14scl,FSS15,FSS16,FSS17prototypical} is one of the application of few-shot learning, especially focused on semantic segmentation. The goal of FSS is to segment the targeted region of the selected category in the query image with their corresponding annotated masks.

The most prevalent approach of FSS is metric-based prototype learning \cite{FSS2PFE}. Referring to the upper part of \cref{fig:first}), a single or multiple class representative prototype vector is generated by the masked average pooling (MAP) \cite{FSS19sg}. A feature processing network segments the target object in the query image leveraging class representative prototype vectors. Many researchers have tried to get more guidance from prototype vectors adopting different mechanism, for example, PANet \cite{FSS18panet}, PFENet \cite{FSS2PFE}, SG-One Net \cite{FSS19sg}, CANet \cite{FSS21canet}, ASGNet \cite{FSS20asg}. However, such prototypical networks can lose detailed spatial information of an image due to masked average pooling operation. In this context, we propose a \textbf{M}ulti-\textbf{S}imilarity and \textbf{A}ttention \textbf{N}etwork (MSANet) consisting of two guiding modules. Referring to the lower part of \cref{fig:first}, the network includes a multi-layer similarity module and an attention module. It is expected that two modules will support prototype learning paradigms and guide the MSANet to fine segmentation. 

Recent works have represented that FSS networks can be upgraded by utilizing visual correspondences \cite{visualcor} of support images and query images. To establish a more meaningful correspondence, dense intermediate layers \cite{dense1,dense2,dense3} and correlation tensor learning \cite{correspondence1,correspondence2,correspondence3} techniques are adopted. Juhong Min \etal designed HSNet \cite{FSS9Hsnet} that suggested a hyper-correlation squeeze network with the multi-layer dense feature correlation-based on 4D tensors. In addition to this, we propose a multi-similarity module that extracts multi-layer feature correlation from a backbone network and applies a simple convolution block to the feature. We also propose a lightweight CNN attention block for paying more attention to the target class content of an image. Following the architecture of BAM \cite{BAM}, we employ a base learner and an ensemble module to refine the segmentation results. We summarize our primary contribution to the FSS challenge as follows:

\begin{itemize}
    \item We propose a multi-layer similarity module to get an informative visual correspondence between a support image and a query image.  
    \item We propose a simple but effective attention module leveraging support images and their corresponding masks to better understand the class-relevant information.
    \item The MSANet outperforms existing FSS networks and shows the state-of-the-art (SOTA) results on PASCAL-$5^i$ \cite{FSS1shaban} and COCO-20\textsuperscript i \cite{coco20} FSS benchmarks under 1-shot and 5-shot settings.
\end{itemize}

\section{Related Work}
\label{sec:rel}
\textbf{Semantic Segmentation:} Semantic segmentation is one of the computer vision tasks to classify each pixel on a given image within specified categories \cite{sem1,sem2,sem3,sem4}. Thanks to advances in fully convolutional networks (FCNs) \cite{sem3}, many model structures such as encoder-decoder-based UNet \cite{unet}, Pyramid Pooling Module (PPM) based PSPNet \cite{PSP} and an Atrous Spatial Pyramid Pooling (ASPP) based deeplab \cite{deeplab} have been proposed for improving segmentation performance. Moreover, a series of vision techniques are suggested, including dilated convolution \cite{dilated}, multi-level feature aggregation \cite{multipath} and attention mechanism \cite{sem-attention}. However, conventional segmentation models require a sufficient amount of annotated data and are difficult to predict unseen categories without fine-tuning, thus hindering practical application to some extent. 

\textbf{Few-shot Learning:}
To tackle these issues, FSL is introduced with the aim of understanding unseen categories with only a few annotated samples. FSL approaches can be further subdivided into three branches: (i) optimization-based \cite{FSL2,optimization1,optimization2}, (ii) augmentation-based \cite{augmen1,augmen2}, and (iii) metric-based \cite{FSS17prototypical,metric1,metric-relation}. The optimization-based methods suggest gradient update strategies to overcome data bias and improve the generalization of the model. The augmentation-based methods address the lack of data by generating synthetic training images. Our work is closely related to the metric-based methods that aim to learn a general metric function to compute the distances between a query image and a support image. There have been outstanding advancements in these metric-based methods. As one of them, matching networks \cite{FSL3} utilize a special kind of mini-batches called episodes to match training and testing environments. Relation networks \cite{metric-relation} convert query and support images to 1x1 vectors and then perform classification based on the Cosine Similarity (CS). Furthermore, prototypical networks \cite{FSS17prototypical}, which directly leverage the feature representations (i.e., prototypes) computed through global average pooling operation, are proposed. 

\textbf{Few-shot Segmentation:}
Shaban, \etal \cite{FSS1shaban} proposed OSLSM, one of the pioneering works of FSS, to generate classifier weights for query image segmentation. The first branch took support images as input and produced a vector of parameters, and the second branch took these parameters as well as query images and generated a segmentation mask as an output. Afterward, the prototype learning paradigm \cite{FSS17prototypical} was introduced for better information extraction from a support image and a query image. SG-One \cite{FSS19sg} introduced masked average pooling operation for computing class representative prototype vectors, yielding the spatial similarity map. CANet \cite{FSS21canet} proposed two dense comparison networks with an iterative refine module. PFENet \cite{FSS2PFE} calculated the CS on high-level features without trainable parameters to create a prior mask and introduced a feature enrichment module. Instead of prototype expansions, ASGNet \cite{FSS20asg} offered a superpixel-guided clustering approach to extract multiple prototypes from the support image, and used an allocation strategy to reconstruct the support feature-map. However, most of the prototype learning methods can lead to spatial structural loss. To fully exploit the features of foreground objects, there is room for improvement in using the class representative prototype vectors.
On the other hand, finding visual correspondences and processing correlation tensors show prominent results in FSS \cite{dense1,dense2,FSS9Hsnet}.
HSNet \cite{FSS9Hsnet} was trained to squeeze a dense feature correlation tensor and transform it into a segmentation mask via high-dimensional convolutions. However, high-dimensional convolutions (4D convolutions) have high spatial and time complexity. To extract a lightweight CNN feature, DENet \cite{denet} introduced a guided attention module to estimate the weights of novel classifiers inspired by traditional attention mechanisms. Tao Hu \etal \cite{tao} proposed an attention-based multi-context guiding network that fuses small-to-large scale context information to guide query branches globally. Instead of working on feature extraction or visual correspondences, BAM \cite{BAM} introduced a new way for FSS, which uses an extra block of the supervised model trained on base classes. The supervised model predicts the base classes from the query image and helps the meta learner to suppress false predictions. Motivated by recent advances in a visual correspondence and an attention mechanism, we propose a multi-layer similarity module and a lightweight attention module in the context of prototypical networks to take FSS networks to the next level.

\section{Problem description}
FSS aims to train a model with base classes and segment novel classes from query images with a few annotated support samples. Current approaches typically train FSS models called a meta learner within a meta-learning paradigm, known as episodic training \cite{FSL3}. Given two image sets \textit{D}\textsubscript{train} (base classes) and \textit{D}\textsubscript{test} (novel classes), the models are expected to learn transferable knowledge on \textit{D}\textsubscript{train} (base classes) with sufficient annotated samples. They have exhibited good generalization capability on \textit{D}\textsubscript{test} (novel classes) with a very few annotated examples. In particular, both sets are composed of numerous episodes, each with a small support set $S=\{(x_{s(i)},m_{s(i)})\}_{i=1}^k$ and a query set $Q=\{(x_{q},m_{q})\}$, where $x^*$ and $m^*$ represent a raw image and its corresponding binary mask for a specific category, respectively. The models are optimized during each training episode to make predictions on the query image $x_q$ under the condition of the support set $S$. Once the training is complete, we will evaluate the performance on \textit{D}\textsubscript{test} across all the test episodes, without further optimization. Like the BAM \cite{BAM}, we follow the same traditional supervised training method for a base leaner network. 
\begin{figure*}[ht]
\centering
\includegraphics[scale=0.1]{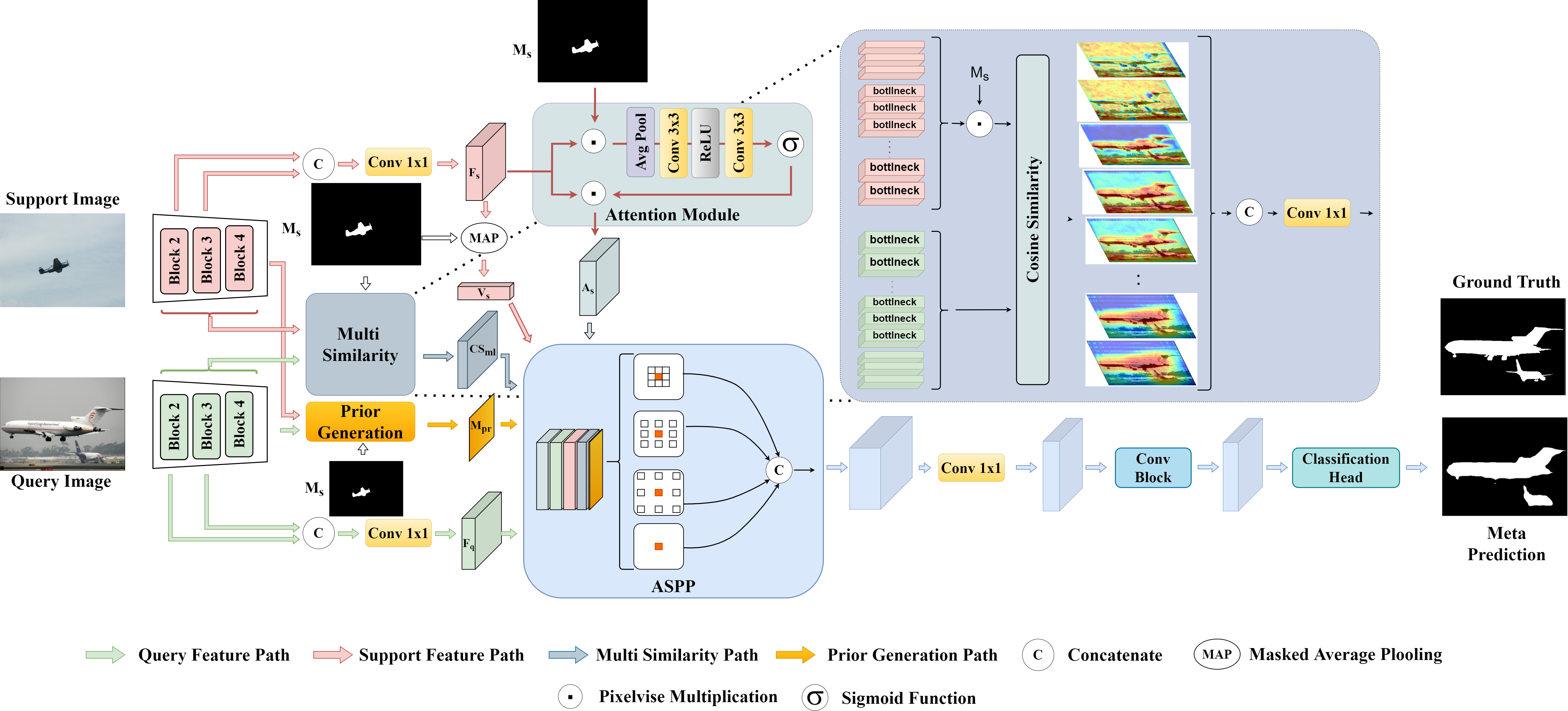}
\caption{ \textbf{Meta Learner Architecture: } Detailed visualization of the meta network for MSANet consisting of the multi-similarity module, the attention module, and the feature processing in the ASPP.}
\label{fig:second}
\end{figure*}

\begin{figure}[t]
\centering
\includegraphics[scale=0.3]{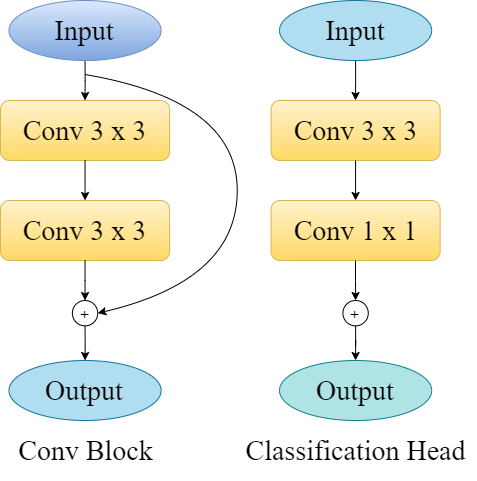}
\caption{The structure for conv block and classification head.}
\label{fig:convblock}
\end{figure}
\section{Proposed Method} 
We propose two guiding modules, the multi-similarity module and the attention module. The former module finds a visual correspondence between the support image and query image, while the latter instructs the FSS network to focus more on the targeted objects of the query image. Taking advantage of a visual correspondence and an attention mechanism, we assist the prototypical network to get more accurate segmentation results.
\begin{figure*}[t]
\centering
\includegraphics[scale=0.12]{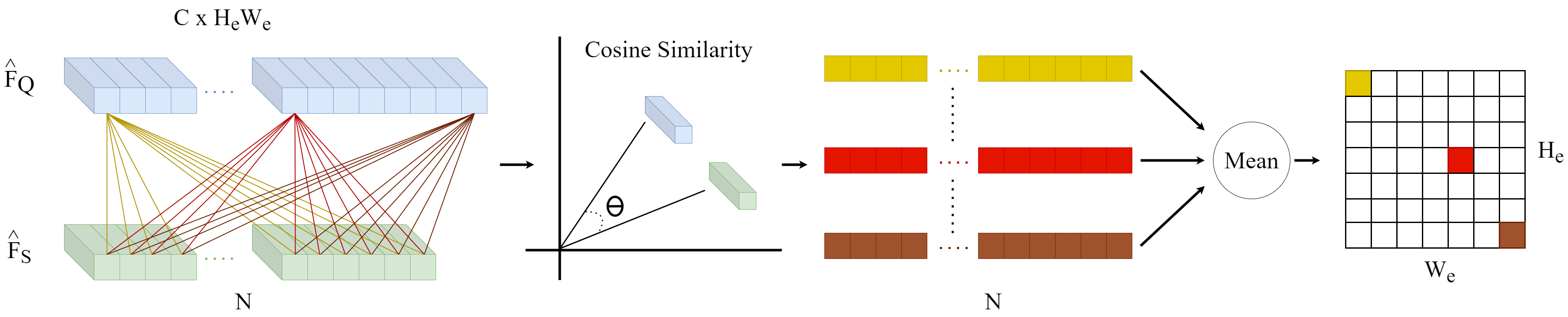}
\caption{The process of computing a visual correspondence.}
\label{fig:CS}
\end{figure*}

\textbf{Model Architecture:}
\cref{fig:second} shows the architecture of the MSANet. First, the features of the query image and the support image are extracted from a pre-trained backbone network. The support features extracted from block 2 and 3 and their corresponding masks are utilized to find a class representative prototype vector $V_s$. These features and their mask are fed to the attention module for finding the attention feature-map. The attention module first masks the support feature and then uses a simple convolutional network to produce a foreground-focused attention feature-map. The query feature and support feature generated from block 4 are utilized to generate a prior mask $M_{pr}$ following \cite{FSS2PFE}. At the same time, all features of the query image and the support image extracted from block 2, 3, and 4 are exploited to generate visual correspondences by leveraging the multi-similarity module. In the module, the CS distances between multi-layers query features and support features are calculated, and simple 1 $\times$ 1 $Conv$ is applied to the features. Details for this module are mentioned in \cref{multi-similarity}. The generated visual correspondence, attention map, prior mask, and prototype vector along with query features are fed to the feature enrichment ASPP module. To focus on the approximate information of features, the dilated version of the ASPP module is utilized. After obtaining rich features from the ASPP module, a simple convolution block is used for feature processing. The classifier head consisting of 3 $\times$ 3 $Conv$ and 1 $\times$ 1 $Conv$ is utilized to produce a binary meta prediction mask. The structure of the convolution block and the classifier head is illustrated in \cref{fig:convblock}. Finally, the output of the meta learner is refined with a base learner \footnote{PSPNet trained on base classes} using an ensemble module.

\subsection{Multi-Similarity Module}\label{multi-similarity}
In this module, a pair of query image ($ I_q $) and support image ($ I_s $), such as $(I_q,I_s)$ $\in$ $\mathbb{R} ^{3\times H \times W}$, are input to the backbone network\footnote{VGG16,ResNet50,ResNet101}. The backbone network pretrained with base classes is frozen during the training process for generalization on unseen categories. To compute the visual correspondence, the last three blocks of the backbone network remain the same spatial size. We extract the last three block feature-maps of the query image as $\hat{F_Q}$ using \cref{eq:1} and the support image as $\hat{F_S}$ with dimension of $ \mathbb{R} ^{C^{b}\times H_\epsilon \times W_\epsilon} $ using \cref{eq:2}, where $C^b$ represents channel size according to bottleneck $b$ and $\epsilon$ represents an image size, respectively. 
\begin{equation}
  \hat{F_Q}= \{ ( F_q^{b_n,b} )_{b_n=0}^{B_N}\}_{b=2}^B
  \label{eq:1}
\end{equation}
\begin{equation}
  \hat{F_S}= \{ ( F_s^{b_n,b} )_{b_n=0}^{B_N}\}_{b=2}^B
  \label{eq:2}
\end{equation}
Here, $B$ represents the block number and $B_N$ represents the bottleneck of $B$ block , respectively. For instance, $F_q^{1,2}$ represents the query feature extracted from the first bottleneck of block 2. Each support feature-map $F_s^{b_n,b}$ is masked with the bi-linear interpolated corresponding mask $M_s$ $\in$  $\{0,1\}^{H\times W}$ using \cref{eq:3} to suppress the activation of background region. By masking the support feature-map, the query feature only correlates with the foreground region of the support image.
 \begin{equation}
  F_{ms}^{b_n,b}= F_s^{b_n,b} \odot \zeta_{\epsilon}(M_s)
  \label{eq:3}
\end{equation}
Here, $\zeta_{\epsilon}(\cdot)$ represents the bi-linear interpolation function that interpolates the support mask $M_s$ $\in$ $\{0,1\}^{H\times W}$ according to the spatial dimension $\epsilon$ followed by the expansion along channel wise such as $\zeta_{\epsilon} : \mathbb{R} ^{H \times W}$ $\Rightarrow$ $\mathbb{R} ^{C^{b}\times H_\epsilon \times W_\epsilon} $, and $\odot$ represents the \textbf{H}adamard product. \\To escape from the over-fitting and to reduce the computation cost, we squeeze the masked support feature-maps $F_{ms}^{b_n,b}$ (\cref{eq:4}) by filtering the mean pixel values such that their dimensions reduce from  $\mathbb{R}^{C^{b}\times H_\epsilon W_\epsilon}  \Rightarrow  \mathbb{R} ^{C^{b}\times N} $, where $ N \ll H_\epsilon W_\epsilon$. The squeezing equation is as follow.

 \begin{equation}
  F_{ms}^{b_n,b,c}=  F_{ms}^{b_n,b} \quad \text{if},  \; [F_{ms}^{b_n,b}\; > c ]
  \label{eq:4}
\end{equation}
Here, $c$ is the $mean$ value of $F_{ms}^{b_n,b}$. To generate a visual correspondence, we first compute pixel-wise cosine distance between squeeze feature-map of the support image $F_{ms}^{b_n,b,c}$ and the extracted feature-map of the query image $F_{q}^{b_n,b}$, following \cref{eq:5}.

\begin{equation}
\begin{split}
  {CS}(x_q,x_s)= mean \left(\phi \{ \frac {x_q^T\cdot x_s } {\parallel x_q \parallel \parallel x_s \parallel} \}\right)  \\
  \quad q \in (1,2,...H_{\epsilon}W_{\epsilon} ) ,
  s \in (1,2,...N )
  \label{eq:5}
 \end{split}
\end{equation}
Here, $ x_q \in F_q^{b_n,b} $, $ x_s \in F_{ms}^{b_n,b,c} $, $\phi$ represents the $ReLU$ function used for the normalization of CS distance tensor and $N$ represents the number of element in $F_{ms}^{b_n,b,c}$, respectively. In \cref{eq:5}, for the first value of $q$, we estimate a cosine distance vector utilizing all values of $N$ and find its mean value to get a single value CS. This computation process repeats for all the values of $q$ to generate a CS map, $ {CS}(x_q,x_s) \in   \mathbb{R} ^{H_{\epsilon} \times W_{\epsilon}}$, as shown in \cref{fig:CS}. The CS map represents the accurate visual correspondence of a single query feature-map with a single support feature-map. The same procedure proceeds for all the extracted feature layers of the query image and the support image to obtain multi-layer visual correspondences using \cref{eq:6}.
\begin{equation}
  CS_{ml}(x_q,x_s)=\left\{CS(x_q,x_s)\right\}_{l=1}^L
 \label{eq:6}
\end{equation}
Here, $L$ is the order number of feature-maps extracted from the backbone network\footnote{L=7,13,30 for VGG16, ResNet50 and ResNet101, respectively}. After finding multi-layer CS, we concatenate them and pass through $1 \times 1$ $Conv$ such as $\mathbb{R}^{C^L \times  H_\epsilon \times W_\epsilon} \Rightarrow \mathbb{R} ^{C^\alpha \times H_\epsilon \times W_\epsilon}$. We choose $\alpha$\ =64, the number of filters for $1 \times 1$ $Conv$. 

\subsection{Attention Module}
In view of the limited number of data provided by novel classes, the information on novel classes may be suppressed by the base classes. To address this issue, we propose a lightweight attention module, which extracts the class-relevant information from the few support samples and directs the network to focus on the targeted region, as shown in \cref{fig:second}. We first extract an intermediate feature-map of the support image and the query image from a backbone network, concatenate them, and apply $1 \times 1$ $Conv$ for dimensionality reduction according to \cref{eq:7}.
\begin{equation}
  F_s^{23}= C_{1 \times1}\{ F_s^2 \; \textcircled{c} \; F_s^3 \}
  \label{eq:7}
\end{equation}
Here, $F_s^2$, $F_s^3$ represent support feature-maps of block 2 and block 3, respectively. These features along with support mask $M_s$ are utilized to get the attention vector using \cref{eq:8}.
\begin{equation}
  V_a=  \sigma(C_{N}(P(F_s^{23} \; \odot \zeta(M_s))))
  \label{eq:8}
\end{equation}
Here, $P$ represents pooling operation, $C_N$ is a convolutional network and $\sigma$ is an activation function, respectively. Finally, a class representative attention feature-map is generated by exploiting the attention vector ($V_\alpha$) (\cref{eq:9}).
\begin{equation}
  A_s=   F_s^{23} \; \odot V_a,
  \label{eq:9}
\end{equation}

\textbf{ASPP and Classifier:} After finding a visual corresponding through the multi-similarity module and an attention feature-map from the attention module, we concatenate them with a prior mask, a class representative prototype vector and intermediate query feature-map. These concatenated features are proceeded through the ASPP module, where a dilated convolution is used for feature enhancement, as shown in \cref{fig:second}. Finally, we apply the convolution block followed by a classifier to the final prediction mask $p_m$. 
\begin{equation}
  p_m=  Softmax(D_m(CS_{ml},A_s,M_{pr},V_s,F_q^{23}))
  \label{eq:10}
\end{equation}
Here, $CS_{ml}$, $A_s$, $M_{pr}$, $V_s$ represent multi-layer similarity, attention features, prior mask, and prototype vector, respectively. $F_q^{23}$ shows the concatenated query features extracted from block 2 and 3 of backbone network. $D_m$ collectively refers to the ASPP, convolution block and classifier.

\textbf{Training Loss:}
The model is trained using a binary cross entropy (BCE) loss. The BCE loss between prediction mask $p_m$ of the query image and its corresponding ground truth mask $m_q$ is calculated.
\begin{equation}
  L_m= \frac{1}{ep}\sum_{i=1}^{ep} BCE(p_{m(i)},m_{q(i)}),
  \label{eq:10}
\end{equation}
Here, $ep$ is the total number of training episodes in each batch. Following the BAM, we also utilize the base leaner loss and the ensemble module loss for end-to-end training.

\textbf{K-shot Segmentation:}
In the K-shot ($K>1$) setting, there are more than one annotated support image. Different approaches have been proposed for K-shot segmentation. Prototype-based networks \cite{FSS2PFE,FSS19sg,FSS17prototypical} mostly took average of the $K$ class representative prototype vectors and then utilized the averaged features to guide the subsequent segmentation process. Whereas, the visual correspondences-based models \cite{FSS9Hsnet} performed $K$ time forward pass and got prediction mask using threshold-based method. In this work, for K-shot segmentation, we perform $K$ forward pass and compute $K$ time CS $\{CS(x_q,x_s)\}_{l=1}^L$, and then the generated $K$ time CS along layer-wise is averaged. Afterwards, the mean CS $\{CS(x_q,x_s)\}_{l=1}^L$ is propagated to the ASPP module. We take the average of $K$ times generated $A_s$,$V_s$ and $M_{pr}$, respectively. Finally, we utilize the adjustment factor with two fully-connected layers following \cite{BAM}.

\begin{table*}[t]
\centering
\begin{adjustbox}{width=1\textwidth}
\small
\begin{tabular}{l r| c c c c c c| c c c c c c}
\midrule
\multirow{2}{*}{Backbone} & \multirow{2}{*}{Method} & \multicolumn{6}{c|}{1-shot} & \multicolumn{6}{c}{5-shot}\\
&&Fold-0 & Fold-1 & Fold-2 & Fold-3 & MIoU\% & FB-IoU\% & Fold-0 & Fold-1 & Fold-2 & Fold-3 & MIoU\% & FB-IoU\%\\\midrule
\multirow{9}{*}{VGG16} & SG-One (TCYB-19) \cite{FSS19sg} & 40.20 & 58.40 & 48.40 & 38.40 & 46.30 & - & 41.9 & 58.60 & 48.60 & 39.40 & 47.10 & -\\
&PANet (ICCV-19) \cite{FSS18panet} & 42.30 & 58.00 & 51.10 & 41.20 & 48.10 & - & 51.80 & 64.60 & 59.80 & 46.50 & 55.70 & -\\
&FWB (ICCV-19) \cite{coco20} & 47.00 & 59.60 & 52.60 & 48.30 & 51.90 & - & 50.90 & 62.90 & 56.50 & 50.10 & 55.10 & -\\
&CRNet (CVPR-20)\cite{FSS6crnet} & - & - & - & - & 55.20 & - & - & - & - & - & 58.50 & -\\
&PFENet (TPAMI-20) \cite{FSS2PFE} & 56.9 & 68.2 & 54.40 & 52.40 & 58.00 & 72.00 & 59.00 & 69.10 & 54.80 & 52.90 & 59.00 & 72.3\\
&HSNet (ICCV-21) \cite{FSS9Hsnet} & 59.6 & 65.7 & 59.60 & 54.00 & 59.70 & 73.40 & 64.90 & 69.00 & 64.10 & 58.60 & 64.10 & 76.60\\
&BAM(CVPR-22) \cite{BAM} & \underline{63.18} & \underline{70.77} & \underline{66.14} & \underline{57.53} & \underline{64.41} & \underline{77.26} & \underline{67.36} & \underline{73.05} & 70.61 & \underline{64.00} & 68.76 & \textbf{81.10}\\
\cline{2-14}
&Meta Learner & 60.92 & 70.00 & 65.82 & 57.39 & 63.53 & 74.61& 66.82 & 72.05 & \underline{72.41} & 63.90 & \underline{68.80} & 79.62\\
\rowcolor{lightgray}
\cellcolor[HTML]{FFFFFF} &Final & \textbf{64.87} & \textbf{71.47} & \textbf{67.40} & \textbf{59.33} & \textbf{65.76} & \textbf{78.01} & \textbf{69.33} & \textbf{73.51} & \textbf{73.59} & \textbf{65.18} & \textbf{70.40} & \underline{80.50}\\\midrule
\multirow{11}{*}{ResNet50} & PANet(ICCV-19) \cite{FSS18panet} & 44.00 & 57.50 & 50.8 & 44.0 & 49.10 & - & 55.30 & 67.20 & 61.30 & 53.20 & 59.30 & -\\
&CANet (ICCV-19) \cite{FSS21canet} & 52.50 & 65.90 & 51.30 & 51.90 & 55.40 & - & 55.50 & 67.80 & 51.90 & 53.20 & 57.10 & -\\
&PGNet (ICCV-19) \cite{FSS15} & 56.00 & 66.90 & 50.60 & 50.40 & 56.00 & 69.90 & 57.70 & 68.70 & 52.90 & 54.60 & 58.50 & 70.50\\
&CRNet (CVPR-20) \cite{FSS6crnet} & - & - & - & - & 55.70 & - & - & - & - & - & 58.80 & -\\
&PPNet (ECCV-20) \cite{FSS7} & 48.58 & 60.58 & 55.71 & 46.47 & 52.84 & 69.19 & 58.85 & 68.28 & 66.77 & 57.98 & 62.97 & 75.76\\
&PFENet (TPAMI-20) \cite{FSS2PFE} & 61.70 & 69.50 & 55.40 & 56.30 & 60.80 & 73.30 & 63.10 & 70.70 & 55.80 & 57.90 & 61.90 & 73.90\\
&HSNet (ICCV-21) \cite{FSS9Hsnet} & 64.30 & 70.70 & 60.30 & 60.50 & 64.00 & 76.70 & 70.30 & 73.20 & 67.40 & 67.10 & 69.50 & 80.60\\
&VAT (arXiv-21) \cite{vat} & 67.60 & 71.20 & 62.30 & 60.10 & 65.30 & 77.40 & 72.40 & 73.60 & 68.60 & 65.70 & 70.00 & 80.90\\
&BAM (CVPR-22) \cite{BAM} & \underline{68.97} & \underline{73.59} & \underline{67.55} &\underline{ 61.13} & \underline{67.81} & \underline{79.71} & \underline{70.59} & \underline{75.05} & 70.79 & \underline{67.20} & \underline{70.91} & \underline{82.18}\\
\cline{2-14}
&Meta Learner & 63.35 & 70.77 & 65.25 & 59.53 & 64.73 & 75.97 & 70.14 & 74.99 & \underline{71.39} & 66.64 & 70.79 & 81.09\\
\rowcolor{lightgray}
\cellcolor[HTML]{FFFFFF} &Final & \textbf{69.25} & \textbf{74.60} & \textbf{67.84} & \textbf{62.40} & \textbf{68.52} & \textbf{80.44}& \textbf{72.70} & \textbf{76.26} & \textbf{73.52} & \textbf{67.94} & \textbf{72.60} & \textbf{83.23}\\\midrule
\multirow{10}{*}{ResNet101} & FWB (ICCV-19)\cite{coco20} & 51.30 & 64.50 & 56.70 & 52.20 & 56.20 & - & 54.80 & 67.40 & 62.20 & 55.30 & 59.90 & -\\
&PPNet (ECCV-20) \cite{FSS7} & 52.70 & 62.80 & 57.40 & 47.70 & 55.20 & 70.90 & 60.30 & 70.00 & 69.40 & 60.70 & 65.1 & 77.5\\
&DAN (ECCV-20)\cite{FSS11} & 54.70 & 68.60 & 57.80 & 51.60 & 58.20 & 71.90 & 57.90 & 69.00 & 60.10 & 54.90 & 60.50 & 72.30\\
&RePRI (CVPR-21) \cite{repri} & 59.60 & 68.60 & 62.20 & 47.20 & 59.40 & - & 66.20 & 71.40 & 67.00 & 57.70 & 65.60 & -\\
&PFENet (TPAMI’20) \cite{FSS2PFE} & 60.50 & 69.40 & 54.40 & 55.90 & 60.10 & 72.90 & 62.80 & 70.40 & 54.90 & 57.60 & 61.40 & 73.50\\
&HSNet (ICCV’21) \cite{FSS9Hsnet} & 67.30 & 72.30 & 62.00 & 63.10 & 66.20 & 77.60 & 71.80 & 74.40 & 67.00 & 68.30 & 70.40 & 80.60\\
&CyCTR (NIPs-21) \cite{cyctr} & \underline{69.30} & 72.70 & 56.50 & 58.60 & 64.30 & 72.90 & \underline{73.50} & 74.00 & 58.60 & 60.20 & 66.60 & 75.00\\
&VAT (arXiv-21) \cite{vat} & 68.40 & 72.50 & 64.80 & \underline{64.20} & \underline{67.50} & \underline{78.80} & 73.30 & 75.20 & 68.40 & \underline{69.50} & 71.60 & \underline{82.00}\\\cline{2-14}
&Meta Learner & 67.56 & \underline{72.90} & \underline{64.94} & 61.91 & 66.82 & 77.31 & 72.14 & \underline{76.66} & \underline{70.77} & 69.27 & \underline{72.21} & 81.94\\
\rowcolor{lightgray}
\cellcolor[HTML]{FFFFFF} &Final & \textbf{70.80} & \textbf{75.20} & \textbf{67.25} & \textbf{}{64.28} & \textbf{69.13} & \textbf{80.38} & \textbf{73.78} & \textbf{77.84} & \textbf{73.14} & \textbf{71.20} & \textbf{73.99} & \textbf{84.30}\\\midrule\end{tabular}
\end{adjustbox}
\caption{\label{pascal} Comparison of the MSANet with other FSS networks on PASCAL-$5^i$ under 1-shot and 5-shot settings. The results with \underline{underlined} denote the second best and with \textbf{bold} shows best performance. The row of the meta learner represents the prediction result for the MSANet without the base learner and the ensemble module.}
\end{table*}

\begin{table*}[h]
\centering
\begin{adjustbox}{width=0.8\textwidth}
\small
\begin{tabular}{l r| c c c c c | c c c c c }\midrule
\multirow{2}{*}{Backbone} & \multirow{2}{*}{Method} & \multicolumn{5}{c|}{1-shot} & \multicolumn{5}{c}{5-shot}\\
&&Fold-0 & Fold-1 & Fold-2 & Fold-3 & MIoU\% & Fold-0 & Fold-1 & Fold-2 & Fold-3 & MIoU\% \\\midrule
\multirow{11}{*}{ResNet50} & HFA (TIP-21) \cite{FSS4} & 28.65 & 36.02 & 30.16 & 33.28 & 32.03 & 32.69 & 42.12 & 30.35 & 36.19 & 35.34 \\
& ASGNet (CVPR-21) \cite{FSS20asg}& - & - & - & - & 34.56 & - & - & - & - & 42.48\\
& RePRI (CVPR-21) \cite{repri} & 32.00 & 38.70 & 32.70 & 33.10 & 34.10 & 39.30 & 45.40 & 39.70 & 41.80 & 41.60 \\
& PPNet (ECCV-20) \cite{FSS7} & 28.10 & 30.80 & 29.50 & 27.70 & 29.00 & 39.00 & 40.80 & 37.10 & 37.30 & 38.50 \\
& PFENet (TPAMI-20) \cite{FSS2PFE}& 36.50 & 38.60 & 34.50 & 33.80 & 35.80 & 36.50 & 43.30 & 37.80 & 38.40 & 39.00\\
& HSNet (ICCV-21) \cite{FSS9Hsnet}& 36.30 & 43.10 & 38.70 & 38.7 & 39.20 & 43.30 & 51.30 & 48.20 & 45.00 & 46.90 \\
& VAT (arXiv-21) \cite{vat}& 39.00 & 43.80 & 42.60 & 39.70 & 41.30 & 44.10 & 51.10 & 50.20 & 46.10 & 47.90\\
& CyCTR (NIPs-21) \cite{cyctr}& 38.90 & 43.00 & 39.60 & 39.80 & 40.30 & 41.10 & 48.90 & 45.20 & 47.00 & 45.60 \\
& BAM (CVPR-22) \cite{BAM} & \underline{43.41} & \underline{50.59} & \textbf{47.49} & 43.42 & \underline{46.23} & 49.26 & 54.20 & \underline{51.63} & \underline{49.55} & 51.16 \\
\cline{2-12}
&Meta Learner & 42.35 & 48.60 & 42.99 & \underline{43.97} & 44.48 & \underline{49.35} & \underline{58.31} & 50.40 & 49.19 & \underline{51.81}\\
\rowcolor{lightgray}
\cellcolor[HTML]{FFFFFF} & Final & \textbf{45.72} & \textbf{54.05} & \underline{45.92} & \textbf{46.44}& \textbf{48.03} & \textbf{50.30} & \textbf{60.89} & \textbf{53.00} & \textbf{50.47} & \textbf{53.67}\\\midrule
\multirow{6}{*}{ResNet101} & FWB (ICCV-19) \cite{coco20}& 17.00 & 18.00 & 21.00 & 28.90 & 21.20 & 19.10 & 21.50 & 23.90 & 30.10 & 23.70 \\
& DAN (ECCV-20) \cite{FSS11} & - & - & - & - & 24.40 & - & - & - & - &29.60 \\
& PFENet (TPAMI-20) \cite{FSS2PFE} & 36.80 & 41.80 & 38.70 & 36.70 & 38.50 & 40.40 & 46.80 & 43.20 & 40.50 & 42.70 \\
& HSNet (ICCV-21) \cite{FSS9Hsnet} & 37.20 & 44.10 & 42.40 & 41.30 & 41.20 & 45.90 & 53.00 & 51.80 & 47.10 & 49.50\\
\cline{2-12}
&Meta Learner & \underline{43.89} & \underline{51.98} & \underline{45.51} & \underline{47.55} & \underline{47.23} & \underline{50.49} & \underline{59.41} & \underline{54.31} & \underline{53.70} & \underline{54.48}\\
\rowcolor{lightgray}
\cellcolor[HTML]{FFFFFF} & Final & \textbf{47.83} & \textbf{57.43} & \textbf{48.65} & \textbf{50.45}& \textbf{51.09} & \textbf{53.23} & \textbf{62.25} & \textbf{55.43} & \textbf{56.30} & \textbf{56.80}\\\midrule\end{tabular}
\end{adjustbox}
\caption{\label{cocotable}Comparison of the MSANet with other FSS networks on COCO-$20^i$ under 1-shot and 5-shot settings. The results with \underline{underlined} denote the second best and with \textbf{bold} shows best performance. The row of the meta learner represents the prediction result for the MSANet without the base learner and the ensemble module.}
\end{table*}

\begin{figure*}[t]
\centering
\includegraphics[scale=0.19]{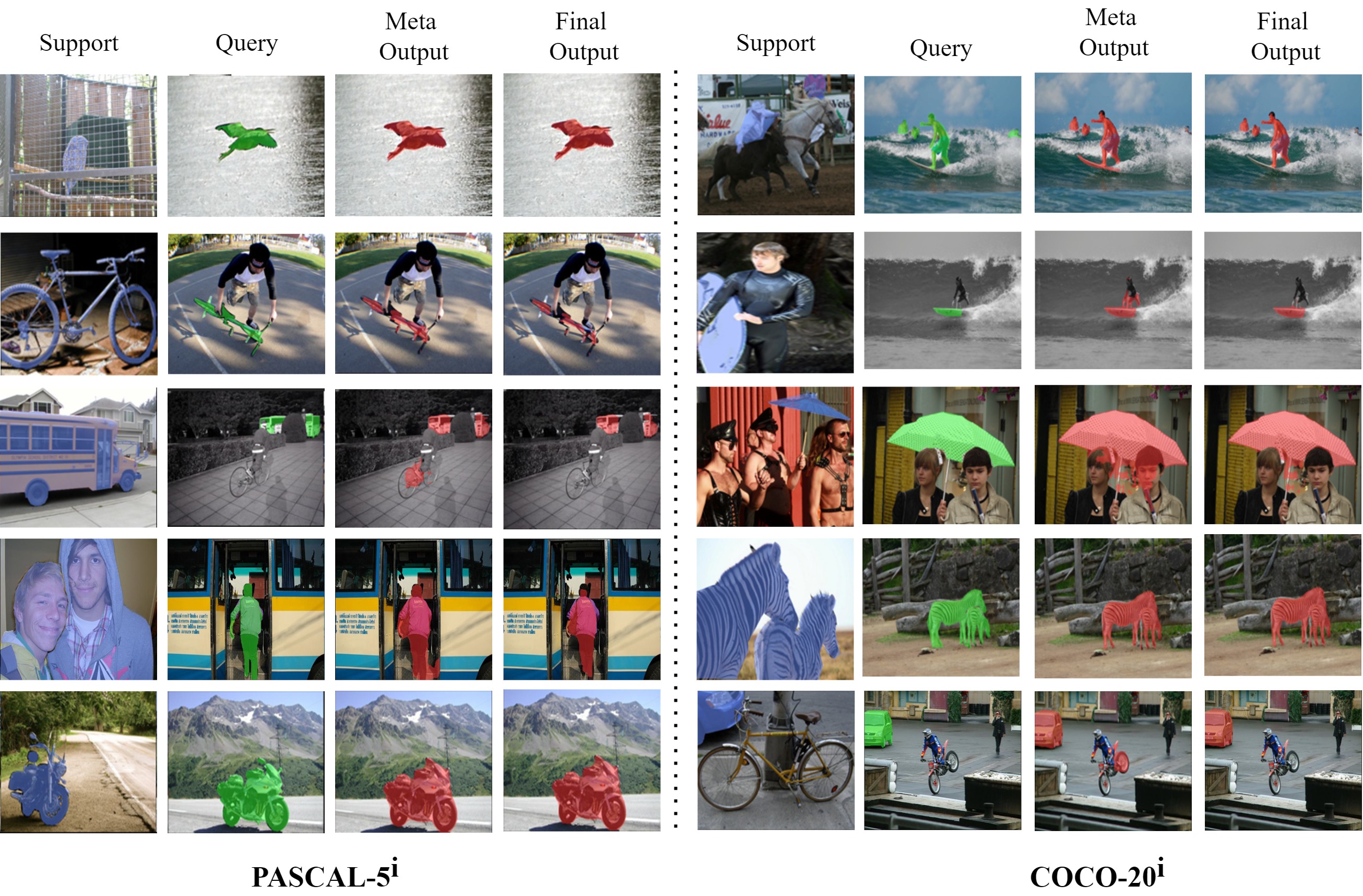}
\caption{The examples of the prediction results for the MSANet on PASCAL-$5^i$ and COCO-$20^i$ under 1-shot setting. The support images with ground-truth masks (blue), the query images with GT masks (green), the meta results (red), and the final results (red) are represented in each row, from left to right. The column of the meta output represents the prediction results of the MSANet without the base learner and the ensemble module.}
\label{fig:visual}
\end{figure*}
\section{Experiments}
\subsection{Implementation Setup}
In this section, three backbone networks\footnote{VGG16 \cite{vgg},ResNet50 \cite{resnet},ResNet101 \cite{resnet}} are used for PASCAL-$5^i$ \cite{FSS1shaban} dataset and two backbone networks\footnote{ResNet50,ResNet101} are used for COCO-$20^i$ \cite{coco20}. We adopted two-way training \cite{BAM}, where the base learner is trained using the supervised protocol. The meta-learner is trained using the traditional episodic training paradigm \cite{FSS17prototypical}. We use the same base-learner as in BAM and fix the parameters during meta learner training. Here, we employ the stochastic gradient descent optimizer with learning rate 5e-2 for 200 epochs on PASCAL-$5^i$ and 50 epochs on COCO-$20^i$, respectively. In both datasets, the batch size is set to 8, and the data augmentation techniques described in \cite{FSS2PFE} are applied. To limit the impact of selected support-query image pairs on performance, we calculate the average results of 5 runs with varied random seeds. The training of the MSANet is implemented in the PyTorch environment, running on the NVIDIA A100 40GB server.

\textbf{Benchmark Dataset:}
We evaluate the performance of the MSANet on standard benchmark datasets, PASCAL-$5^i$ and COCO-$20^i$. PASCAL-$5^i$ consists of 20 object classes generated from PASCAL VOC 2012 \cite{pascal} with additional annotations from SDS \cite{SDS}. COCO-$20^i$ consists of 80 object classes compiled from MSCOCO \cite{coco}. The object categories are equally distributed into 4-folds such as $\{5^i:i \in \{0,1,2,3\}\}$ for PASCAL-$5^i$ , $\{20^i:i \in \{0,1,2,3\}\}$ for  COCO-$20^i$, respectively. Models are trained on 3 folds and tested on the remaining one fold based on a cross-validation protocol. The validation fold consists of 1000 random pairs of support images and query images.

\textbf{Evaluation Metric}
We employ mean intersection over-union (mIoU) and foreground-background IoU (FBIoU) as the assessment metrics, following prior FSS approaches \cite{FSS2PFE,FSS21canet,BAM,FSS9Hsnet}. 
\begin{equation}
  mIoU= \frac{1}{C} \sum_{c=1}^C IoU_c 
  \label{eq:11}
\end{equation}
\begin{equation}
  FB_{IoU}= \frac{1}{2} (IoU_f + IoU_b )
  \label{eq:12}
\end{equation}
In \cref{eq:11}, $C$ and $IoU_c$ represent total classes in the targeted fold and the intersection over union of class $c$, respectively. In \cref{eq:12}, $IoU_f$ and $IoU_b$ represent foreground and background intersection over union values in the targeted fold, respectively.
\subsection{Result Analysis}
We compare the performance of the MSANet with the other FSS networks using PASCAL-$5^i$ and COCO-$20^i$ datasets. The experiments are conducted with different backbone networks in 1-shot and 5-shot scenarios. The performances of the MSANet are verified in both quantitative and qualitative paradigms. 

\textbf{Quantitative Results}:
\cref{pascal} and \cref{cocotable} illustrate the performances of the MSANet along with other FSS approaches. In both FSS dataset benchmarks, PASCAL-$5^i$ and COCO-$20^i$, the MSANet outperforms all prior FSS networks under 1-shot and 5-shot settings in term of ${mIoU}$ and $FB_{IoU}$. Compared to SOTA \cite{BAM}, for PASCAL-$5^i$ benchmark, in 1-shot setting, the MSANet with VGG16, ResNet50, and ResNet101 backbones show performance improvements of 1.35\%, 0.71\%, and 1.63\%, respectively, and in 5-shot setting, of 1.64\%, 1.69\%, and 2.39\%, respectively. For COCO-$20^i$ benchmark, the networks with ResNet50 and ResNet101 backbones outperform with high margin such as 1.8\% and 2.5\% (1-shot) and 9.89\% and 7.3\% (5-shot), respectively. 

\textbf{Qualitative Results:}
\cref{fig:visual} presents the examples of the prediction results of the MSANet under 1-shot setting for PASCAL-$5^i$ and COCO-$20^i$. In the figure, first two columns, third column, and the forth column represent the examples of support images and the query images, the output of the meta part for the MSANet, and the output of the MSANet, respectively. As shown in \cref{fig:visual}, it is found that the predicted results of the MSANet are almost identical to the ground truth in pixel wise segmentation, which demonstrate the performance of the MSANet. 

\begin{table}[ht]
\centering
\begin{adjustbox}{width=0.5\textwidth}
\small
\begin{tabular}{c c c c| c | c}
\midrule
Multi Sim & Prototype & Attention & Prior Mask & Meta mIoU(\%) & Final mIoU(\%) \\
\midrule-& \Checkmark & \Checkmark & \Checkmark & 65.12 & 67.47 \\
\Checkmark &- & \Checkmark & \Checkmark & 65.84 & 69.04 \\
\Checkmark & \Checkmark &- & \Checkmark & 66.54 & 68.50 \\
\Checkmark & \Checkmark &\Checkmark &- & 66.28 & 68.71 \\
\midrule
\Checkmark& & \Checkmark& & 65.25 & 68.87 \\
\Checkmark & \Checkmark & \Checkmark & \Checkmark & \textbf{66.82} & \textbf{69.13} \\
\midrule
\end{tabular}
\end{adjustbox}
\caption{\label{module-wise} The result of the ablation study. The meta mIoU represents the prediction of the MSANet without the base learner and the ensemble module. }
\end{table}

\begin{figure}[t]
\centering
\includegraphics[scale=0.18]{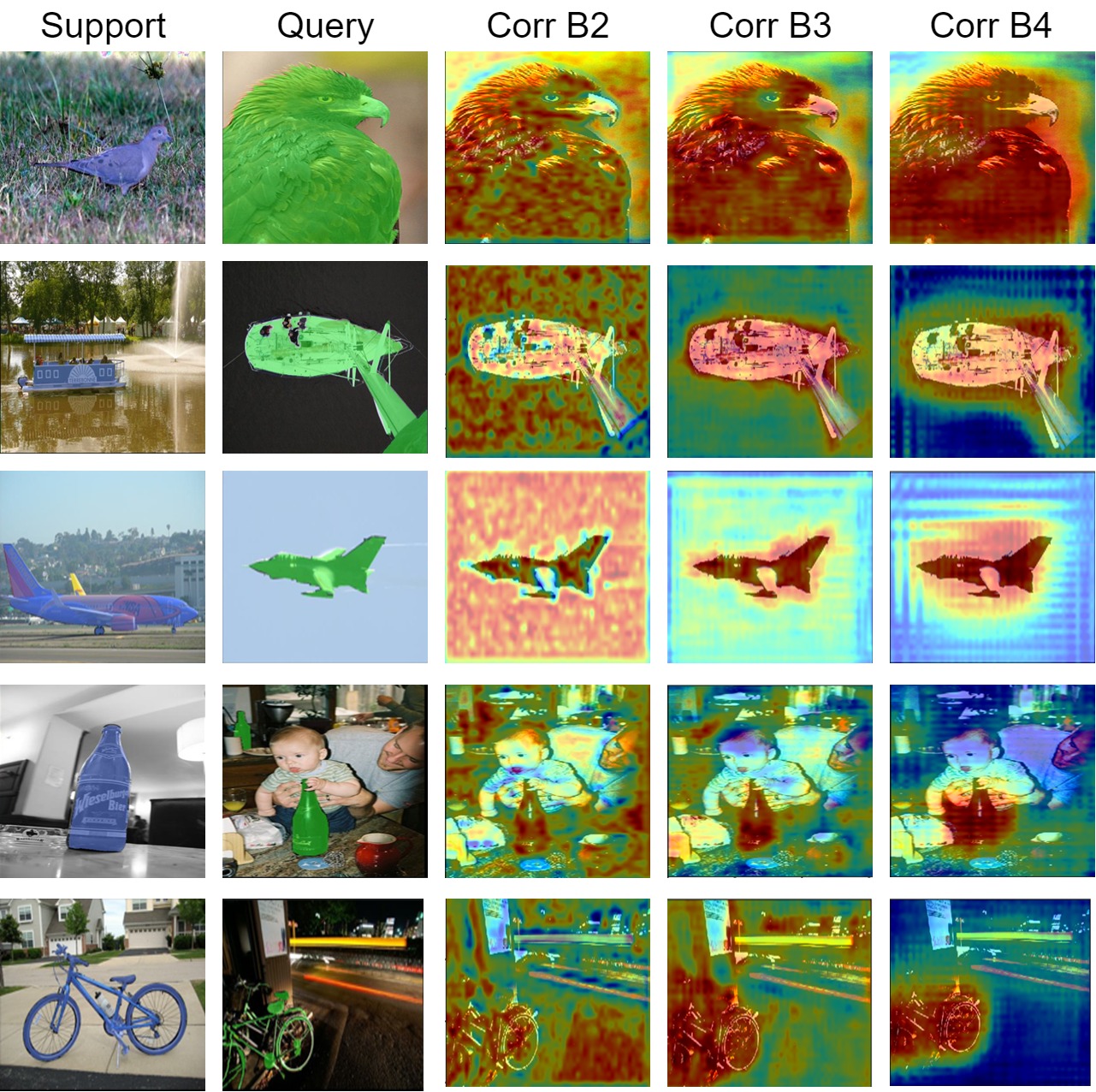}
\caption{Visualization of multi-layer similarity correlation from different blocks. CorrB2, CorrB3 and CorrB4 represent the multi-similarity from block 2, block 3, and block 4 of the backbone network, respectively}
\label{fig:correlation}
\end{figure}

\subsection{Ablation Tests}
We undertake a series of ablation tests using ResNet101 backbone on PASCAL-$5^i$ under 1-shot setting. This test can evaluate the impact of each component on segmentation performance and verify its effectiveness.

\textbf{Performance of Module:} 
\cref{module-wise} shows the effectiveness of each module in the MSANet through the ablation tests. Compared to the performance of the MSANet, the network without the multi-similarity, the attention, prototype, and prior mask module descends it to 1.66\%, 0.63\%, 0.1\%, and 0.42\%, respectively. These results demonstrate that two proposed modules, multi-similarity and attention, have more impact on performance improvement than the previous FSS prototype approaches (prior mask, prototype vector). The fifth row of \cref{module-wise} shows that the network with only two modules achieves 68.87\%, which is higher than all previous FSS performance shown in \cref{pascal}. Referring to the final row of \cref{module-wise}, the combination of the two modules and the previous FSS prototype modules leads to the MSANet accomplishing the highest performance. The table also shows that the base learner and the ensemble modules play a significant role in the MSANet.

\textbf{Layer Selection for Multi-Similarity:}
To understand the impact of each feature layer in computing similarity correlation, we experiment with different blocks of backbone networks. In the MSANet, multi-similarity correlations are computed using the three blocks from the backbone. \cref{fig:correlation} exhibits the visualization of multi-similarity correlation according to different blocks with an energy map representing the average value of all similarities in one block. The correlation with low-level features holds the detailed information but lacks the objectness. On the contrary, the images with high-level features can understand the approximate information but loses the details such as edges. Accordingly, low-level (block 2), mid-level (block 3), and high-level features (block 4) are used for the computation of semantic similarity to obtain diverse context information about target objects. We figure out that leveraging visual correspondence by combining multiple feature layers of a backbone network can provide more guidance in segmenting target objects. 

\begin{figure}
\includegraphics[scale=0.23]{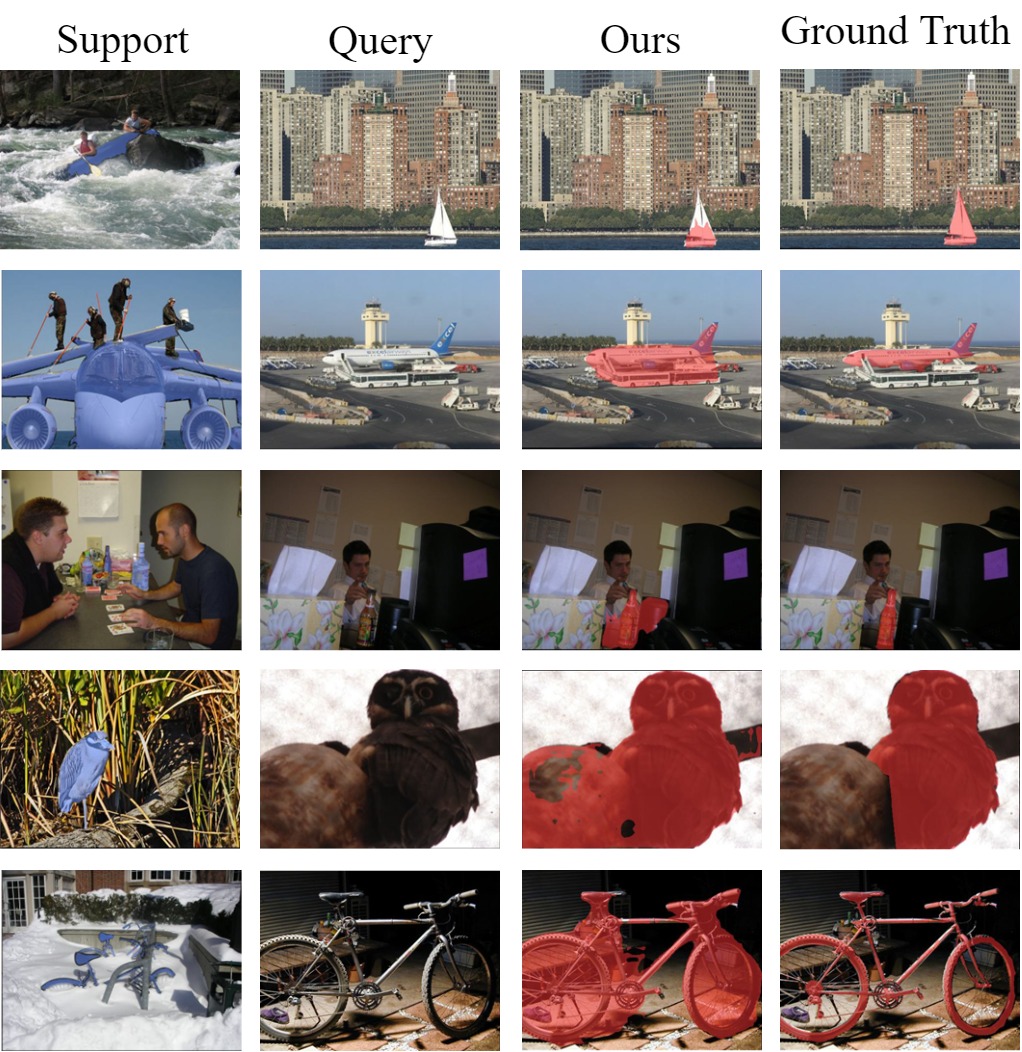}
\caption{Visualization of failure cases.}
\label{fig:failure}
\end{figure}

\textbf{Failure Case Study:}
We visualize the failure cases of the MSANet in \cref{fig:failure}. The predicted results of the MSANet are sometimes unclear and discontinuous, possibly due to the model's failure to obtain accurate clues from the support images. These issues similarly appear in few-shot semantic segmentation tasks, and are still one of the challenges in the computer vision field. The results in \cref{fig:failure} imply that failure cases may be proportional to the complexity of a pair of support and query image. In addition, input pairs that are relatively lacking in visual representation can result in inaccurate segmentation masks. These difficulties in FSS can suggest future work directions.

\section{Conclusion}
In this paper, we propose the MSANet for few-shot image segmentation. Two new modules, named multi-similarity and attention, are introduced to the FSS to overcome the shortcomings of existing prototype-based models. The first module exploits the multiple feature-maps of the support images and the query images to generate an informative visual correspondence between them. The second module helps the MSANet to concentrate more on class-relevant information. Extensive experiments and ablation studies prove the effectiveness of the proposed network. We success to achieve the SOTA performances for 4-benchmark datasets, PASCAL-$5^i$ and COCO-$20^i$ datasets under 1-shot and 5-shot settings, respectively.

{\small
\bibliographystyle{ieee_fullname}
\bibliography{egbib}
}

\end{document}